\documentclass[conference,10pt]{IEEEtran}
\usepackage{epsfig,rotating,setspace,latexsym,amsmath,epsf,amssymb,amsfonts,bm,theorem,subfigure,epstopdf}
\usepackage{cite,authblk}
\usepackage{bbm}
\usepackage{color}
\usepackage{mathtools}
\usepackage{algorithm}
\usepackage{algpseudocode}
\usepackage{verbatim}
\usepackage{xcolor}

\algrenewcommand\algorithmicforall{\textbf{foreach}}
\algrenewcommand\algorithmicindent{.8em}

\algnewcommand\algorithmicforeach{\textbf{for each}}
\algdef{S}[FOR]{ForEach}[1]{\algorithmicforeach\ #1\ \algorithmicdo}

\IEEEoverridecommandlockouts
\allowdisplaybreaks

\begin{document}

\title{Multi-Modal Semantic Communication}

\author{Matin Mortaheb \qquad Erciyes Karakaya \qquad Sennur Ulukus\\
        \normalsize Department of Electrical and Computer Engineering\\
        \normalsize University of Maryland, College Park, MD 20742\\
        \normalsize \emph{mortaheb@umd.edu} \qquad \emph{rerciyes@umd.edu} \qquad \emph{ulukus@umd.edu}}

\maketitle

\begin{abstract}
Semantic communication aims to transmit information most relevant to a task rather than raw data, offering significant gains in communication efficiency for applications such as telepresence, augmented reality, and remote sensing. Recent transformer-based approaches have used self-attention maps to identify informative regions within images, but they often struggle in complex scenes with multiple objects, where self-attention lacks explicit task guidance. To address this, we propose a novel \emph{Multi-Modal Semantic Communication} framework that integrates text-based user queries to guide the information extraction process. Our proposed system employs a cross-modal attention mechanism that fuses visual features with language embeddings to produce soft relevance scores over the visual data. Based on these scores and the instantaneous channel bandwidth, we use an algorithm to transmit image patches at adaptive resolutions using independently trained encoder-decoder pairs, with total bitrate matching the channel capacity. At the receiver, the patches are reconstructed and combined to preserve task-critical information. This flexible and goal-driven design enables efficient semantic communication in complex and bandwidth-constrained environments. 
\end{abstract}

\section{Introduction}

Semantic communication has emerged as a promising paradigm for next-generation communication systems, aiming to prioritize the transmission of information that is most relevant to a given task rather than transmitting raw data \cite{sagduyu2023task,sagduyu2023vulnerabilities,mortaheb2024transformer}. In particular, semantic communication over vision modalities seeks to encode and transmit only the most informative visual content necessary for the receiver to complete a desired task, significantly reducing communication costs while maintaining task performance. Such approaches hold strong potential for applications including bandwidth-constrained telepresence, augmented reality, and remote sensing, where efficient and goal-aligned transmission of visual information is critical.

To identify and transmit the most informative parts of an image for a given task, recent works have leveraged transformer-based architectures \cite{dosovitskiy2020image} for semantic communication by exploiting self-attention scores \cite{vaswani2017attention}. Notably, \cite{mortaheb2024efficient,mortaheb2024transformer} introduce a transformer-aided semantic communication system that uses the internal self-attention maps of a vision transformer as a semantic encoder to prioritize spatial regions most relevant to the task during encoding and transmission. This method achieves substantial gains in communication efficiency, particularly for images with spatially sparse semantic content.
However, approaches that rely solely on self-attention mechanisms have several key limitations. First, in complex visual scenes involving multiple objects, background clutter, or ambiguous semantic cues, self-attention alone may fail to consistently isolate the regions most pertinent to the communication objective. In such scenarios, attention maps may be diffused or overly focused on visually salient but semantically irrelevant regions. Moreover, in dynamic settings such as video transmission, the semantic structure of the scene evolves over time, making purely vision-driven attention insufficient to maintain stable, task-consistent prioritization.
Second, these methods require retraining or fine-tuning whenever the communication task changes. For example, shifting the goal from image classification to anomaly detection demands re-optimizing the entire semantic encoder in \cite{mortaheb2024efficient,mortaheb2024transformer}. Since the encoder is based on a heavy transformer architecture, this retraining is computationally expensive and time-consuming. This lack of flexibility limits the practicality of self-attention-only frameworks, especially in real-time or multi-purpose communication scenarios where rapid task adaptation is essential.

\begin{figure}[t]
\centerline{\includegraphics[width=1\linewidth]{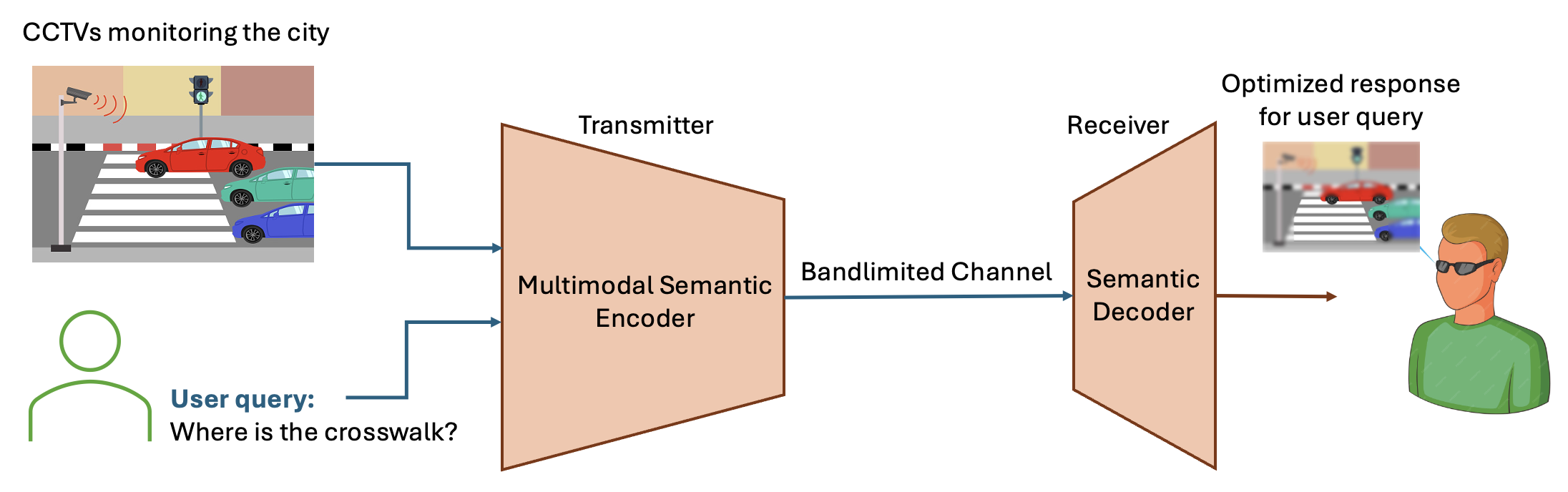}}
  \caption{Multi-modal semantic communication framework.}
\centering
\label{fig:MMSC_framework}
\vspace*{-0.4cm}
\end{figure}

To address these challenges, we introduce a multi-modal semantic communication framework that leverages cross-modal attention between images and user-provided textual queries. This enables user-guided selection of important regions, refining the system’s focus toward semantically relevant content that aligns with the user’s intent rather than relying solely on vision-driven saliency. Using pretrained vision–language models such as CLIP \cite{Ilharco_Open_Clip_2021}, we first extract embeddings for both the image and the textual query. To better capture the visual context, we apply a content-dependent transfer mechanism \cite{jiao2024collaborative} to refine the query embedding based on the image content, ensuring stronger image–text alignment.
Next, leveraging the image embedding, the conditioned text embedding, and a mask proposal mechanism \cite{cheng2021mask2former}, we compute an informativeness score for each image patch, quantifying its relevance to the user query. These scores are integrated into a multi-resolution encoding algorithm, allowing the system to transmit patches at varying quality levels while respecting bandwidth constraints. This approach is particularly effective in images where self-attention alone fails to capture the true semantic priority or where task-driven relevance is not purely visual.
Importantly, our framework does not require retraining or fine-tuning for each new user query. Instead, the system dynamically adapts to the provided instruction, selecting and prioritizing image segments based on their relevance to the query. This greatly improves flexibility and makes the framework suitable for real-time and multi-task communication settings.

Previous work in open-vocabulary semantic segmentation, such as MAFT+ \cite{jiao2024collaborative}, leverages cross-attention mechanisms to fuse image and text representations, enabling pixel-level segmentation for arbitrary text queries. These methods typically generate hard binary masks (0 or 1) indicating the presence or absence of queried concepts in the image. In contrast, our approach computes soft informativeness scores for each image patch, reflecting the degree of relevance to the user query rather than making binary decisions. Furthermore, while open-vocabulary segmentation models often rely on short, single-word queries drawn from predefined datasets, our system supports more expressive, natural-language queries. To the best of our knowledge, this is the first work to use cross-attention-derived multi-modal relevance scores for semantic communication, enabling adaptive transmission of image content based on query relevance.

We evaluate our proposed framework using a custom-made subset of the COCO dataset \cite{lin2014microsoft}. First, we demonstrate that the model can visually identify and extract the regions of an image that are highly relevant to the user query. We then compare the task accuracy achieved using reconstructed data against a baseline approach that relies solely on the self-attention mechanism of a vision transformer to extract semantic information.

The main contributions of this paper are summarized as follows: I) We propose a semantic extraction module that jointly processes an image and a text-based user query, leveraging cross-attention to identify the most informative image regions relevant to the query. II) We use an algorithm that categorizes image patches into multiple resolution levels based on the instantaneous channel rate. Patches with higher informativeness scores are reconstructed with higher fidelity, enabling efficient and task-aware bandwidth usage. III) We evaluate the effectiveness of the proposed multi-modal semantic communication framework across three complementary metrics to verify that it reliably preserves the query-relevant content in the reconstructed images.

\section{System Model} \label{sec:system_model}

In this section, we introduce a multi-modal semantic communication framework that fuses visual and textual modalities to enable semantically guided compression and transmission under bandwidth constraints. The system model consists of an encoder–decoder architecture. The encoder comprises two main components: a multi-modal semantic extractor and a patch-wise multi-resolution encoder. After encoding, the image patches are transmitted over the channel and processed by resolution-specific decoders to reconstruct the final image. Unlike conventional semantic communication systems, our framework operates with two inputs—an image and a user-provided command—which together determine the semantic priorities for transmission (see Fig.~\ref{fig:MMSC_framework}).

\subsubsection{Multi-Modal Semantic Extractor}

\begin{figure*}[]
    \centerline{\includegraphics[width=0.75\linewidth]{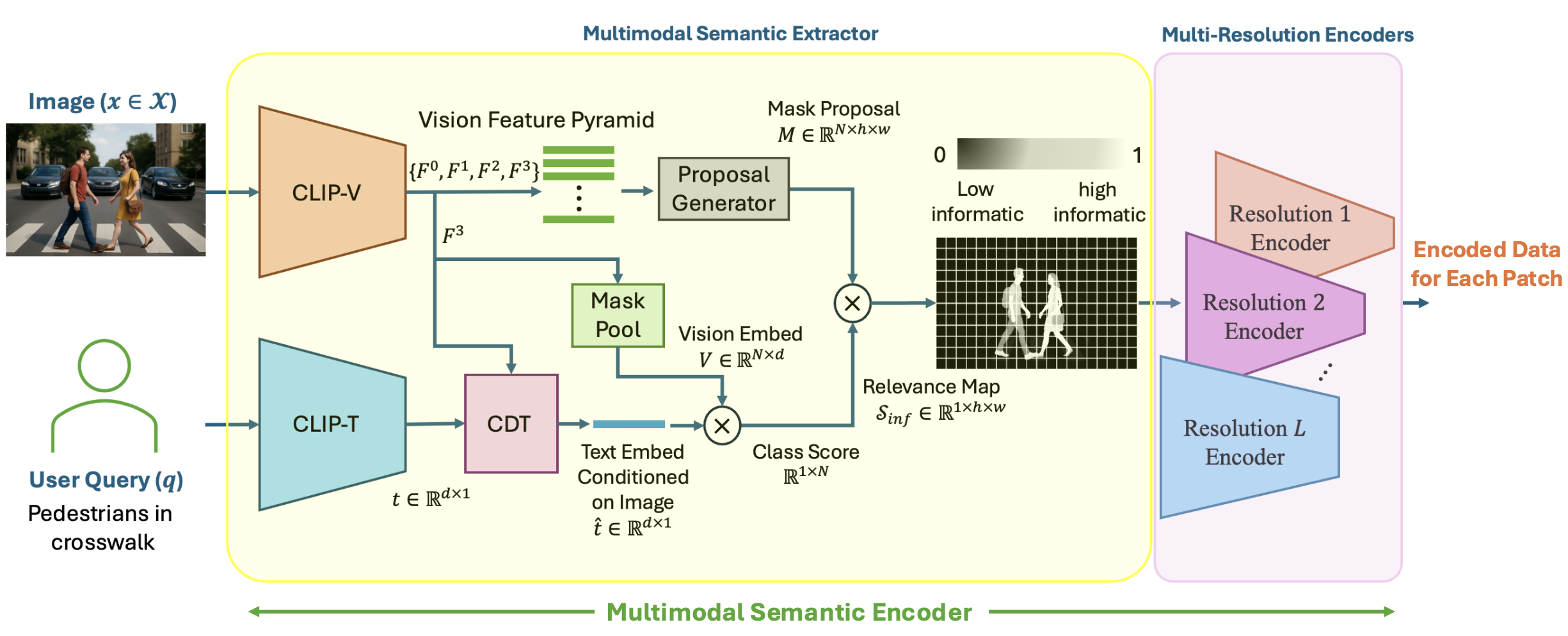}}
    \caption{Multi-modal semantic encoder framework.}
    \centering
    \label{fig:MMSC_encoder}
    \vspace*{-0.3cm}
\end{figure*}

The goal of the multi-modal semantic extractor is to fuse the vision and text inputs to produce a semantic score indicating the most informative regions of an image given the user’s textual query. Our semantic extractor builds upon the MAFT+ framework \cite{jiao2024collaborative}, originally designed for open-vocabulary semantic segmentation (OVSS). However, unlike MAFT+, our objective is not to assign binary class labels to pixels, but to compute a soft relevance map that quantifies how strongly each region aligns with an arbitrary user query. First, as shown in Fig.~\ref{fig:MMSC_encoder}, the image input $x \in \mathbb{R}^{3\times h\times w}$ is processed by a convolutional CLIP-Vision (CLIP-V) backbone to extract a feature pyramid $F = \{F^0, F^1, F^2, F^3\}$, where $F^i \in \mathbb{R}^{d\times h_i \times w_i}$ and the spatial strides relative to the input are $\{4,8,16,32\}$, respectively, with a feature dimension $d$. 

These visual features are passed to a MaskFormer proposal generator \cite{cheng2021mask2former, cheng2021per}, which first upsamples and fuses the multi-scale features into a dense per-pixel embedding map $E_{\mathrm{pixel}} \in \mathbb{R}^{d\times h\times w}$. A fixed set of $N$ learnable query tokens $\{q_i\}_{i=1}^N$ is passed through a transformer decoder that performs cross-attention with $E_{\mathrm{pixel}}$, producing a set of mask embeddings $ E_{\mathrm{mask}} = [E_{\mathrm{mask}}^{(1)}, \dots, E_{\mathrm{mask}}^{(N)}] \in \mathbb{R}^{N \times d}$. These embeddings are then projected back onto the image domain via dot product to compute the mask logits $M \in \mathbb{R}^{N \times h \times w}$, where each mask logit is given by
\begin{align}
    m_i(h, w) = E_{\mathrm{mask}}^{(i)\top} E_{\mathrm{pixel}}[:, h, w].
\end{align}
The resulting mask logits represent unbounded scores indicating the likelihood of each pixel belonging to each proposal. Note that each mask is responsible to find one object/region in the given image in a class-agnostic mannor.

During training, a set of class prompts $C$ indicating all possible objects is embedded via CLIP-Text (CLIP-T) encoder $f_t(\cdot)$, producing text embeddings $T=[t_1,\dots,t_{|C|}]\in\mathbb{R}^{d\times|C|}$. $T$ is refined then via two transformer cross-attention layers called content-dependent transfer (CDT), attending on the flattened highest-level visual feature $F^3_{\mathrm{flat}}\in\mathbb{R}^{d\times h_3w_3}$ 
\begin{align}
    T_{j+1} = T_j + \mathrm{TransLayer}_j(T_j, F^3_{\mathrm{flat}}), \nonumber\\
    \text{ for } j=0,1, \text{ with } T_0 = T.
\end{align}
This would yield a conditioned embedding $\hat T\in\mathbb{R}^{d\times|C|}$. Meanwhile, mask pooling on $F^3$ using each mask logit yields mask embeddings $V\in\mathbb{R}^{N\times d}$. We compute per-mask classification scores
\begin{align}
    S_{\mathrm{cls}}=\left(V\hat T\right) ^\top\in\mathbb{R}^{|C|\times N}.
\end{align}
When projecting these scores onto the full image plane, we form dense semantic maps of size $|C|\times h\times w$ via
\begin{align}
    S(c,h,w)=\sum_{i=1}^N S_{\mathrm{cls}}(c,i)\cdot\sigma(m_i(h,w)),
\end{align}
where $\sigma$ is the sigmoid activation. 

Three losses are used during training. First, the mask proposal loss $L_P$ uses Hungarian matching \cite{kuhn1955hungarian} between the predicted masks $M$ and ground-truth masks, combining binary cross-entropy and Dice losses. This loss updates only the MaskFormer proposal generator, keeping CLIP-V frozen.
Second, to ensure that the model is sensitive to the quality of predicted masks we use mask-aware classification loss that explicitly aligns the CLIP-based classification confidence with the true segmentation quality. The core intuition is that high CLIP similarity scores should correspond to accurate mask predictions with high Intersection over Union (IoU). Let $S_{\mathrm{IoU}}$ represent the ground-truth IoU between the predicted mask and its corresponding annotation of user query. The mask-aware loss enforces consistency between these two scores through a SmoothL1 objective:
\begin{align}
    \mathcal{L}_{\mathrm{ma}} = \mathrm{SmoothL1}\left( S_{\mathrm{cls}},\, S_{\mathrm{IoU}} \right).
\end{align}
By regressing the CLIP similarity score toward the true IoU, the model learns to assign higher confidence to masks with stronger spatial alignment, effectively incorporating mask quality into the learned representation. This mechanism encourages the system to down-weight inaccurate proposals and amplifies the contribution of high-quality ones, ultimately improving both mask ranking and segmentation accuracy.
Third, the representation compensation loss $\mathcal{L}_{rc}$ preserves original CLIP-V representation by matching multi-scale pooled features between the fine-tuned CLIP-V and a frozen CLIP-V* using SmoothL1 loss (detailed explanation is available in \cite{jiao2024collaborative})
\begin{align}
    \mathcal{L}_{rc}=\sum_{k\in\{1,2,4\}}\mathrm{SmoothL1}(F_k^p,\hat F_k^p).
\end{align}
Gradients from $\mathcal{L}_{ma}$ and $\mathcal{L}_{rc}$ are backpropagated into CLIP-V (with $\mathcal{L}_{ma}$ also updating CDT), while CLIP-T remains entirely frozen throughout.

At inference time, unlike training, the framework processes the text input as user command $q$ through the frozen CLIP-T encoder $f_t(\cdot)$ to produce $t\in\mathbb{R}^{d\times 1}$. The generated text embedding $t$ passes through the same refinement pipeline CDT to generate $\hat t$ which is a refined embedding of the text conditioned on the image features. Meanwhile, the image is passed through CLIP-V and MaskFormer to compute the visual features $F$, mask logits $M$, and pooled mask embedding $V$. The pixel-level semantic relevance map is computed as
\begin{align}
    S_{\mathrm{inf}}(h,w)=\sum_{i=1}^N\langle\hat t,v_i\rangle\cdot\sigma(m_i(h,w)).
\end{align}
This yields a relevance map $S_{\mathrm{inf}}\in\mathbb{R}^{1\times h\times w}$, which encodes how well each pixel aligns with the user’s textual query.

While our multi-modal semantic extractor is inspired by OVSS, there are key differences. OVSS typically produces binary segmentation masks indicating whether each object class is present. In contrast, we use a soft relevance score between 0 and 1, representing the confidence that a pixel corresponds to the queried concept. This soft score enables patch-wise variable-resolution encoding, which we detail in the next section. Additionally, in OVSS, the text input often consists of a list of single-word object categories. In our case, the user command is a single free-form query, potentially referencing multiple objects. The objective is to generate a mask that captures the degree to which each region of the image matches the intent of the query.

\subsubsection{Multi-Resolution Encoder/Decoder}

Given the relevance map $S_{\mathrm{inf}}$ and the available bandwidth budget $B$, the input image is first partitioned into $P = (h/p)(w/p)$ non-overlapping patches of size $p \times p$. For each patch $x_i$, we compute a semantic importance score $s_i$ by averaging the per-pixel values within the patch from $S_{\mathrm{inf}}$. Each patch is then assigned to one of $L$ predefined resolution levels, indexed by $\ell_i \in \{1, \dots, L\}$, where each level corresponds to a different encoding bitrate $r_{\ell_i}$. Higher semantic scores are mapped to higher resolution levels (i.e., higher $r_{\ell_i}$) to preserve critical information, while less relevant patches are compressed more aggressively to save bandwidth.
The patch-to-resolution assignment is performed using a resource allocation Algorithm \cite[Algorithm 1]{mortaheb2024efficient}, ensuring that the total bitrate across all patches satisfies the constraint $\sum_{i=1}^P r_{\ell_i} \leq B$.
Once the assignments are finalized, each patch $x_i$ is encoded using its designated encoder $\mathcal{E}_{\ell_i}(x_i)$, transmitted, and then decoded at the receiver via the corresponding decoder $\mathcal{D}_{\ell_i}$ to produce the reconstructed patch $\hat{x}_i$. The final reconstructed image $\hat{x} \in \mathbb{R}^{3 \times H \times W}$ is assembled by placing all $\hat{x}_i$ back into their original positions.

This framework enables \textit{task-driven, semantic-aware communication} by strategically allocating communication resources based on the semantic relevance of image content with respect to the user’s text query. As a result, semantically important regions, those most aligned with the user’s intent, are preserved at higher fidelity, while less relevant regions are encoded more compactly, thus achieving efficient and goal-oriented transmission under bandwidth constraints.

\section{Experimental Result}

We evaluate the proposed method along two key dimensions: reconstruction fidelity and the ability to preserve query-relevant semantic content. Our objective is to determine how effectively the reconstructed image reflects the user’s intent as expressed in the textual query. In particular, we measure how well the system retains semantically important regions that correspond to the queried concept, demonstrating that our attention-guided compression framework can capture user-specific visual information even under strict bandwidth constraints.

\subsection{Dataset Specifications}
For evaluation of the multi-modal semantic communication framework, we use the custom-made COCO validation dataset \cite{lin2014microsoft}, a widely used benchmark in computer vision that contains 5,000 images of complex everyday scenes, annotated for object detection, segmentation, and captioning tasks. From this dataset, we select a subset of 200 images for evaluation and adapt the dataset setup to include user queries. For fair comparison with our benchmark \cite{mortaheb2024efficient}, all images are resized to a resolution of $320 \times 480$ pixels. For each image, we generate a query in one of two ways: (i) by selecting a single object label from the image’s annotated categories, or (ii) by randomly selecting two annotated objects and forming a query in the format “object 1 and object 2.” In both cases, the query is appended to the image sample to simulate a user command. Additionally, the ground-truth segmentation masks corresponding to the queried objects are included to provide reference for evaluating semantic alignment between the query and the reconstructed image.

\subsection{Model and Hyperparameters}

We incorporate MAFT+ \cite{jiao2024collaborative} as the multi-modal semantic extractor. For the CLIP model within MAFT+, we use the ConvNeXt-Large variant from OpenCLIP \cite{Ilharco_Open_Clip_2021}. The embedding dimension is set to $d=1024$. The proposal generator component follows the default configuration of Mask2Former \cite{cheng2021mask2former}, with the number of class-agnostic mask proposals set to $N = 100$. For evaluating semantic relevance, we compute text-image similarity scores using the CLIP ViT-Large-Patch14-336 model for both vision and text encoders.

\subsection{Results and Analysis}

In this framework, we adopt the same five-resolution scheme as before, with encoding sizes of $L = 0, 12, 24, 48, 192$ bytes. The lowest resolution ($L=0$) skips transmission of the patch, while the highest resolution ($L=192$) transmits the raw image patch without compression corresponding to the full $3 \times 8 \times 8$ bytes. The intermediate resolutions use a learned encoder to compress each patch to the desired size, and a decoder to reconstruct it back to the original dimensions. To standardize our setup, we resize all images to dimensions $320 \times 480$. Given a patch size of $8 \times 8$, this results in exactly 2,400 patches per image. Therefore, the total channel bandwidth for transmitting an entire image ranges from 0 (no patches sent) up to $2400\times 192$ bytes, which corresponds to sending all patches at the highest resolution without any compression. This range defines the limits of our system under varying bandwidth constraints.

\begin{figure}[h]
    \centerline{\includegraphics[width=0.9\linewidth]{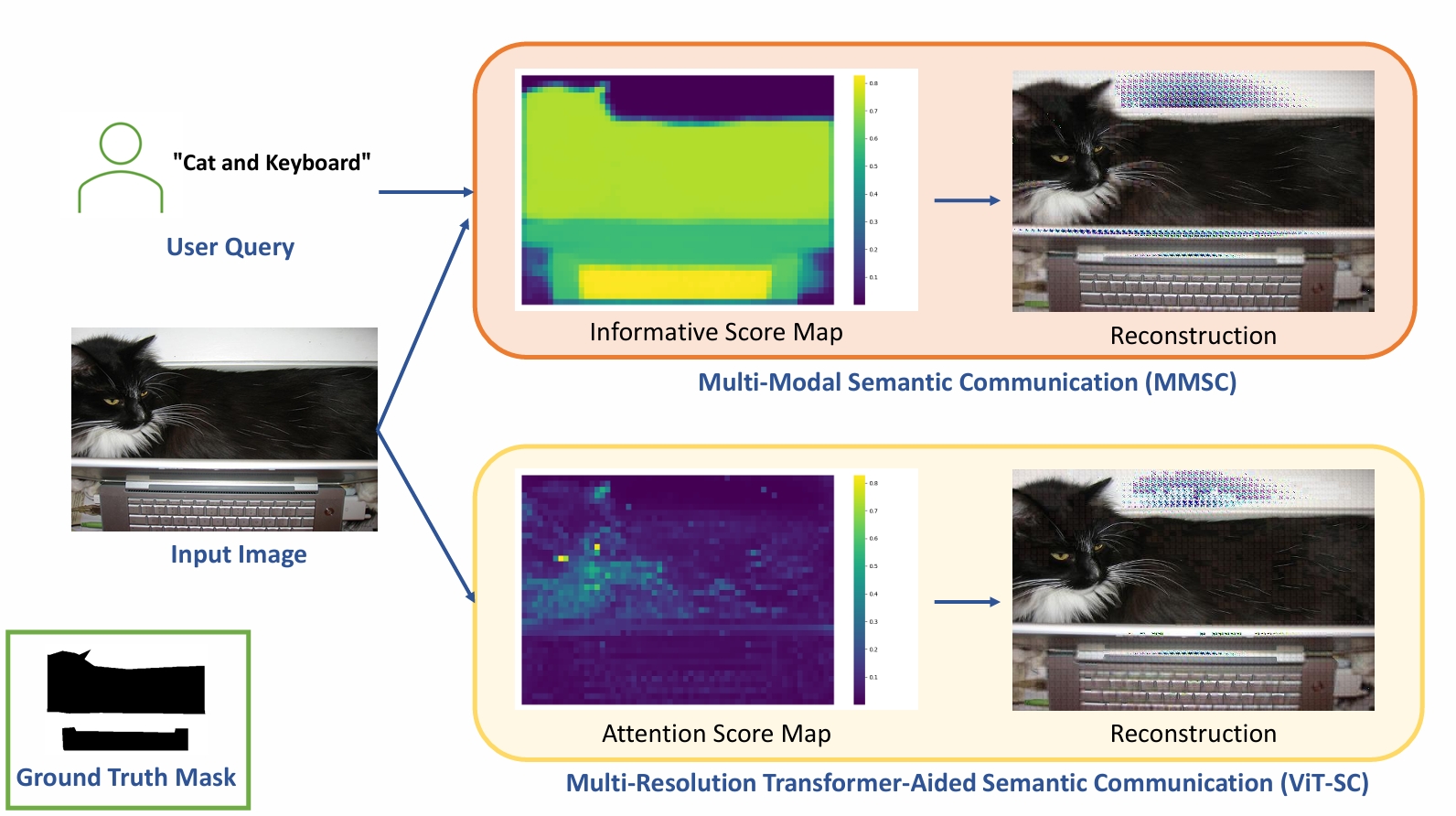}}
    \caption{An example image from the dataset, transmitted and reconstructed using the MMSC and ViT-SC \cite{mortaheb2024efficient} models under a 50\% channel rate constraint.}
    \centering
    \label{fig:MMSC_ViT_example_project3}
    \vspace*{-0.4cm}
\end{figure}

Our framework aims to reconstruct the most informative regions of the image based on a user-provided text query. These regions are identified by a vision-language model and prioritized during encoding to receive higher resolution when bandwidth is limited. Fig.~\ref{fig:MMSC_ViT_example_project3} shows a qualitative comparison between our two semantic communication frameworks: multi-modal semantic communication (MMSC) and multi-resolution transformer-aided semantic communication (ViT-SC\cite{mortaheb2024efficient}). The input image (left) is accompanied by a user query: “Cat and Keyboard”. The ground truth mask highlights the regions corresponding to the queried objects. In the top block, the MMSC framework takes both the image and the user query as input. It generates an informative score map $S_{\mathrm{inf}}$ that identifies semantically relevant regions aligned with the query and reconstructs the image accordingly under a 50\% channel rate. The bottom block shows the output of the ViT-SC framework, which only receives the image as input. It produces an attention score map based on internal attention weights and reconstructs the image under the same bandwidth constraint. As shown, the MMSC framework successfully identifies and reconstructs both queried objects, cat and keyboard, demonstrating strong alignment with the user’s intent. In contrast, the ViT-SC framework primarily focuses on the most salient object (the cat) and fails to consistently highlight or reconstruct the keyboard. This example underscores the advantage of incorporating multi-modal semantic guidance for query-aware communication and adaptive compression. To further evaluate the reconstruction quality and relevance to the user query, we apply three different approaches over 200 image-text pairs we extracted from COCO validation set.

\begin{figure}[h]
    \centerline{\includegraphics[width=0.8\linewidth]{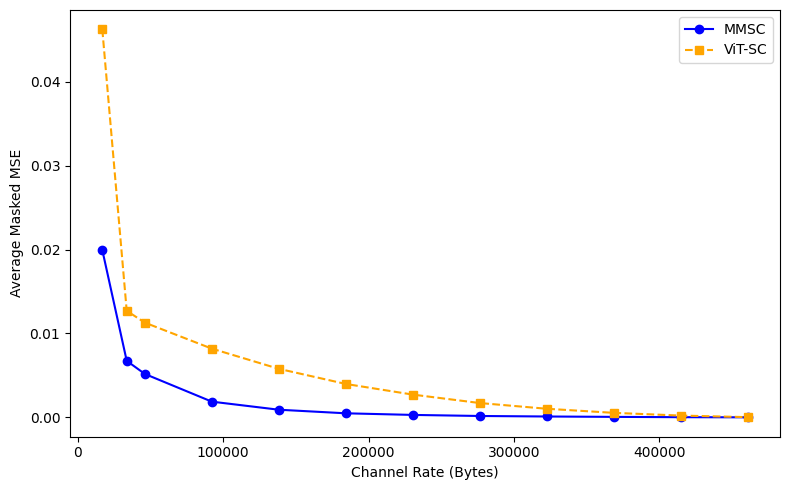}}
    \caption{Average MSE between the original and reconstructed images within the masked area defined by the ground truth segmentation corresponding to the user query.}
    \centering
    \label{fig:Masked_MSE_project3}
    \vspace*{-0.2cm}
\end{figure}

In the first approach, we compute the MSE between the original and reconstructed images, but only within the masked area defined by the ground truth segmentation corresponding to the user query. This region is expected to be reconstructed with the highest fidelity, as it directly relates to the user’s intent. We compare our proposed MMSC framework with the ViT-based framework \cite{mortaheb2024efficient}, which does not incorporate the user query during encoding. For fairness, both frameworks share the same encoder-decoder and resolution control components. Fig.~\ref{fig:Masked_MSE_project3} shows that our framework consistently achieves lower masked MSE across all bandwidth levels, indicating better preservation of the semantically important areas.

\begin{figure}[h]
    \centerline{\includegraphics[width=0.8\linewidth]{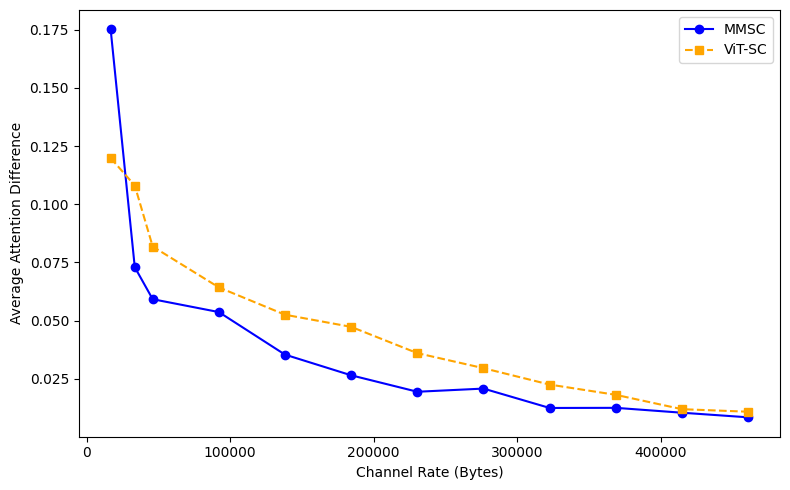}}
    \caption{L1 distance between informative score maps of the original and reconstructed images conditioned on the user query.}
    \centering
    \label{fig:diss_attn_project3}
    \vspace*{-0.3cm}
\end{figure}

In the second approach, we analyze the semantic consistency using the informativeness score $S_{\mathrm{inf}}$, as discussed in Section~\ref{sec:system_model}. We compute $S_{\mathrm{inf}}$ between the original image and text query, and $\hat{S}_{\mathrm{inf}}$ between the reconstructed image and the same text. The L1-distance between these scores reflects how well the informative content is preserved. We calculate this distance for both MMSC and ViT-SC frameworks. As illustrated in Fig.~\ref{fig:diss_attn_project3}, our MMSC framework demonstrates smaller differences between $S_{\mathrm{inf}}$ and $\hat{S}_{\mathrm{inf}}$, outperforming the ViT-based framework across the full range of available bandwidth.

\begin{figure}[h]
    \centerline{\includegraphics[width=0.8\linewidth]{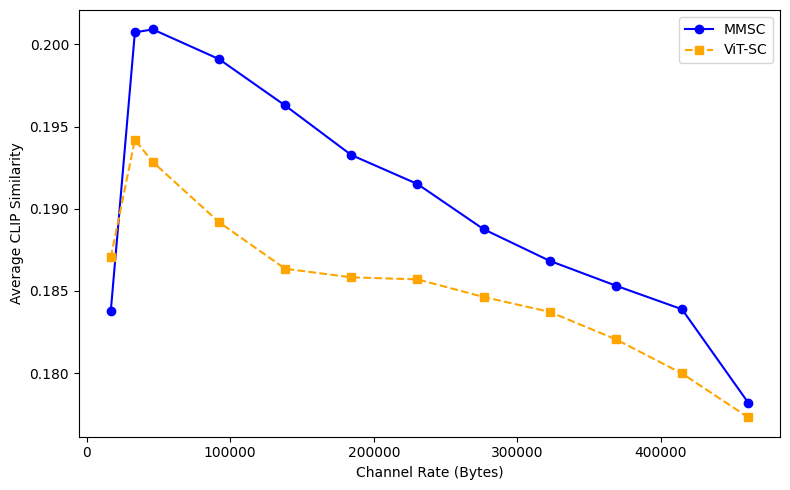}}
    \vspace*{-0.2cm}
    \caption{CLIP relevancy score between reconstructed image and text.}
    \centering
    \label{fig:CLIPScore_project3}
    \vspace*{-0.25cm}
\end{figure}

In the third approach, we evaluate the relevance between the reconstructed image and the input text using the CLIP model. The image is encoded using CLIP-V and the text with CLIP-T, and the dot product of their embeddings provides a relevance score, typically ranging from 0.1 to 0.32. A higher score implies stronger alignment between the visual and textual inputs. We calculate the CLIP score for both our MMSC and ViT-SC reconstructions. Fig.~\ref{fig:CLIPScore_project3} indicates that our framework consistently yields higher CLIP scores, particularly at intermediate bandwidths. This is because, under bandwidth constraints, the model is forced to focus on transmitting only the most relevant patches at higher resolutions. At higher bandwidths, however, additional irrelevant regions may also be transmitted and reconstructed, which can dilute semantic focus and slightly reduce the CLIP score.

\bibliographystyle{unsrt}
\bibliography{reference}
\end{document}